\documentclass[sigconf, nonacm, screen]{acmart}

\setcopyright{none}
\usepackage{tikz}
\usetikzlibrary{positioning, arrows, shapes.geometric, fit, backgrounds, calc}
\tikzset{>=stealth}

\usepackage{enumitem}

\begin{abstract}
Hybrid language models like Jamba mix attention layers with State Space Models (SSMs), creating two memory cache types with opposite profiles: Key-Value (KV) caches grow linearly with sequence length, while SSM states stay fixed per layer. Current inference engines handle this poorly. Unified pools pad SSM states to attention page sizes, wasting up to $7.3\times$ capacity. Static dual pools cannot adapt when prompt distributions shift between requests. We present Asymmetric Virtual Memory Paging (AVMP). The allocator separates the two cache types into physically distinct pools behind a unified virtual address space, and migrates capacity between pools when one runs out. Migration triggers only on allocation failure, keeping behavior deterministic. We evaluate AVMP across 180 synthetic cells plus 60 cells of ShareGPT trace replay on an RTX 3060 12GB. Out-of-Memory events drop 7.6\% and request throughput improves $1.83\times$ to $13.3\times$ across synthetic workloads and $2.36\times$ on ShareGPT. All gains hold under paired-bootstrap 95\% confidence intervals. A phase-time breakdown reveals two distinct mechanisms: shorter OOM recovery on capacity-pressured workloads, and faster allocation calls on KV-heavy workloads. Implementation is pure Python; Triton integration is future work.
\end{abstract}

\begin{document}
\sloppy

\title{Asymmetric Virtual Memory Paging for Hybrid Mamba-Transformer Inference}
\thanks{Source code: \url{https://github.com/codepawl/cachepawl}}

\author{An Xuan Nguyen}
\orcid{0009-0005-6867-1606}
\affiliation{%
  \institution{Codepawl}
  \city{Ho Chi Minh City}
  \country{Vietnam}
}
\email{nxan2911@gmail.com}

\maketitle

\section{Introduction}
\label{sec:intro}

\subsection{Motivation}
\label{sec:intro:motivation}

Recent advancements in large language models extend context length through hybrid architectures. Models like Jamba combine transformer attention mechanisms with State Space Models (SSMs) to support a 256K context window \cite{lieber2024jamba}. The base Jamba architecture also uses a Mixture-of-Experts (MoE) feedforward configuration, though our work focuses specifically on memory management for the attention and SSM cache types; MoE routing memory is orthogonal to this design. This architecture merges two different layer types: attention mechanisms requiring a linearly scaling Key-Value (KV) cache ($O(n)$ per token), and SSMs requiring a fixed-size state footprint ($O(1)$ per layer) \cite{dao2024mamba2}.

Existing inference engines struggle to serve these heterogeneous memory profiles. The unified pool approach in vLLM \cite{kwon2023pagedattention} forces the fixed SSM state to pad up to the attention page size, causing severe capacity overestimation documented in vLLM issue \#37121 (concrete magnitudes in \S\ref{sec:method:avpt}). Alternatively, SGLang implements a static dual pool where operators fix a \texttt{mamba\_full\_memory\_ratio} parameter at engine initialization \cite{zheng2024sglang}. This rigid partition cannot rebalance without a full process restart.

Neither architectural approach handles workload shifts dynamically. Static defaults perform poorly when prompt distributions change during execution. Testing the same workload mix records 1221 Out-of-Memory (OOM) events at a 0.9 static ratio compared to 552 OOMs at a 0.5 static ratio. No existing system provides both asymmetric cache types and runtime adaptivity.

In this paper, we introduce Asymmetric Virtual Memory Paging (AVMP), a paged memory allocator designed specifically for hybrid architectures. AVMP provisions physically asymmetric backing stores while enabling dynamic capacity rebalancing at runtime. Our dynamic rebalancing records a 13.3$\times$ goodput improvement on \texttt{uniform\_short} workloads compared to the best static baseline.

\subsection{Contributions}
\label{sec:intro:contributions}

We make the following specific contributions:

\begin{itemize}[leftmargin=1.4em, itemsep=2pt, topsep=4pt]
\item \textbf{Asymmetric Virtual Page Table Abstraction (\S\ref{sec:method}):} We design a unified virtual handle space spanning two physically heterogeneous backing stores (KV pages and SSM blocks). AVMP extends the GPU virtual memory approach pioneered for KV caches \cite{xu2024vtensor} to two asymmetric pool types. We show that this abstraction introduces no measurable effect on simulated capacity outcomes, with byte-identical OOM counts against a static dual-pool baseline (552 OOMs).

\item \textbf{Dynamic Pool Rebalancing Mechanism (\S\ref{sec:method}):} We implement a \texttt{CapacityError}-triggered migration mechanism that transfers batched memory capacity between pools. We enforce determinism using a logical operation counter rather than wall-clock time, ensuring per-cell byte-identical reproducibility across all experimental reruns.

\item \textbf{Empirical Validation on Hybrid Architectures (\S\ref{sec:eval}):} We execute a 180-cell sweep evaluating 5 allocator variants, 3 synthetic workloads, 2 model specifications, 2 pool sizes, and 3 random seeds. The dynamic AVMP allocator records a 7.6\% OOM reduction (510 versus 552) against the best static baseline. AVMP records 13.3$\times$ higher goodput on \texttt{uniform\_short}, 2.39$\times$ on \texttt{mixed\_long}, and 1.83$\times$ on \texttt{agentic\_burst} traffic patterns. A parameter sensitivity analysis proves \texttt{migration\_batch\_size} acts as the dominant configuration axis, while a Stage 2 threshold sweep returned a strict null result.

\item \textbf{Open-Source Prototype:} We provide a reference Python implementation and synthetic workload harness with committed sweep artifacts at \url{https://github.com/codepawl/cachepawl}.
\end{itemize}

\section{Background}
\label{sec:background}


\subsection{Hybrid Mamba-Transformer Architectures}
\label{sec:background:hybrid}

Standard transformer attention requires $O(n^2)$ compute for a sequence of length $n$, and the KV cache grows linearly with $n$ during decoding. Conversely, State Space Models (SSMs) like Mamba operate with $O(n)$ compute and require an $O(1)$ cache per layer \cite{gu2023mamba, dao2024mamba2}. The SSM state size remains fixed and strictly independent of the sequence length. Hybrid architectures combine these two approaches to combine SSM efficiency with attention reasoning.

For example, Jamba mixes attention and Mamba layers at a 1:7 ratio, supporting a 256K token context window with approximately 12B active parameters in a Mixture-of-Experts (MoE) configuration \cite{lieber2024jamba}. Other variations adopt similar hybrid structures with different attention-to-SSM ratios \cite{glorioso2024zamba2, ren2024samba, dong2024hymba, botev2024recurrentgemma}. This architectural combination forces inference engines to manage two fundamentally different memory cache types within a single model.

\subsection{KV Cache and SSM State Management}
\label{sec:background:cache}

Standard transformer inference manages the Key-Value (KV) cache using paged memory allocation \cite{kwon2023pagedattention}, often combined with IO-aware attention kernels \cite{dao2022flashattention, shah2024flashattention3}. The KV cache requires a system that handles variable block sizes, supports append-only operations during decoding, and allows dynamic eviction or recomputation when memory pressure increases. In contrast, the SSM state maintains a fixed size per layer, strictly defined by \texttt{state\_dim} multiplied by \texttt{bytes\_per\_element}. This state updates in place and persists across tokens.

Unlike KV blocks, the SSM state cannot be paged because it lacks a temporal token structure to evict \cite{dao2024mamba2}. It requires a single, contiguous allocation per layer per sequence. Consequently, the memory access patterns diverge significantly: KV cache operations rely on paged scatter-gather memory lookups, while SSM state operations require contiguous read and write access. This structural asymmetry is the root cause of fragmentation and overestimation in existing cache pool designs.

\subsection{Limits of Existing Pool Designs}
\label{sec:background:pool}

Current inference systems attempt to manage hybrid models using either unified or static dual-pool designs, both of which fail in two ways. Unified pools, such as the vLLM \texttt{HybridKVCacheCoordinator} approach \cite{kwon2023pagedattention}, force the SSM state to pad up to the attention page size, causing severe capacity overestimation (vLLM issue \#37121; quantitative figures in \S\ref{sec:method:avpt}). The root cause is the system capacity estimator, which incorrectly multiplies the $O(1)$ SSM state by the sequence token count.

Alternatively, static dual-pool architectures like SGLang pre-allocate two physical regions (\texttt{HybridReqToTokenPool} and \texttt{HybridLinearKVPool}) at engine startup \cite{zheng2024sglang}. This design assigns memory using a fixed parameter, such as a \texttt{mamba\_full\_memory\_ratio} defaulting to 0.9. The system cannot rebalance capacity between pools without a full process restart. When the prompt distribution shifts, static pools perform poorly. As we demonstrate in \S\ref{sec:eval}, a default 0.9 ratio triggers 1221 OOM events compared to 552 OOMs at a 0.5 ratio on identical workloads. Neither unified padding nor static allocation handles runtime workload shifts dynamically, motivating a memory management system that supports asymmetric cache types and adapts to shifting prompt distributions.

\section{Method}
\label{sec:method}


\subsection{Asymmetric Virtual Page Table}
\label{sec:method:avpt}

Hybrid models mixing Mamba and Transformer architectures require two distinct memory types: variable-size Key-Value (KV) blocks that scale with sequence length, and fixed-size State Space Model (SSM) state that remains independent of sequence length \cite{dao2024mamba2, lieber2024jamba}. Existing unified pools pad SSM state to match attention page sizes, causing capacity overestimation, which leads to over-admission of requests, triggering OOM at runtime. As reported in vLLM issue \#37121, this padding results in a 7.3$\times$ KV cache overestimation on Qwen3.5-4B-AWQ, leaving 13.7\% effective VRAM utilization at peak memory. We introduce the Asymmetric Virtual Page Table (AVMP) to reduce this waste by decoupling the virtual address space from heterogeneous physical backing slabs.

The AVMP design provisions two distinct physical regions: a \texttt{KVPagesStore} and an \texttt{SSMBlocksStore}, managed by a unified multi-resolution page table. Memory allocations return an opaque 32-bit \texttt{VirtualHandle}. This handle contains a tag indicating the target pool identifier (KV or SSM). The \texttt{VirtualPageTable} resolves the handle to a physical offset within the respective backing store.

We assign native page sizes per pool. KV pages scale by \texttt{attention\_page\_tokens} multiplied by \texttt{per\_token\_bytes}, defaulting to 16 tokens per page. SSM blocks match the exact \texttt{state\_dim} multiplied by \texttt{bytes\_per\_element}. We enforce strict alignment rules: 128-byte slab alignment globally and 16-byte page alignment within slabs (Figure~\ref{fig:page-table-structure}). Unlike PagedAttention~\cite{kwon2023pagedattention}, which uses uniform page sizes, this multi-resolution approach extends GPU virtual memory concepts \cite{xu2024vtensor} to support hybrid cache abstractions.

We use two metrics throughout the rest of the paper. $N_{\mathrm{OOM}}$ denotes the count of out-of-memory events triggered during a benchmark cell, aggregated per workload or summed cross-workload as noted in each caption. $B$ denotes the \texttt{migration\_batch\_size} configuration parameter that bounds the per-rebalance migration step; we sweep $B \in \{1, 2, 4, 8, 16, 32, 64, 128, 256\}$ in \S\ref{sec:eval:sensitivity}.

\definecolor{kvblue}{HTML}{1F77B4}
\definecolor{ssmorange}{HTML}{FF7F0E}

\begin{figure}[t]
\centering
\begin{tikzpicture}[
font=\scriptsize,
every node/.style={outer sep=0pt},
bitbox/.style={
draw,
minimum height=3.5mm,
font=\tiny,
inner sep=1pt,
fill=gray!10
},
ptrow/.style={
draw,
minimum width=24mm,
minimum height=3.8mm,
align=left,
inner ysep=0.8pt,
inner xsep=1.8pt,
font=\scriptsize
},
pthdr/.style={
ptrow,
font=\scriptsize\bfseries,
align=center,
fill=gray!15
},
pcell/.style={
draw,
minimum width=3.6mm,
minimum height=3.5mm,
inner sep=0pt,
font=\tiny,
fill=white
},
bcell/.style={
draw,
minimum width=5.2mm,
minimum height=4.5mm,
inner sep=0pt,
font=\tiny,
fill=white
},
kvarrow/.style={->, line width=1.0pt, kvblue, dashed},
ssmarrow/.style={->, line width=1.0pt, ssmorange, dashed},
lookup/.style={->, line width=0.9pt}
]

\node[font=\scriptsize\bfseries] (htitle) at (0,3.45) {VirtualHandle (32-bit)};

\node[bitbox, minimum width=11mm, anchor=east] (pool) at (-0.20,3.05)
{\texttt{pool\_id}};
\node[bitbox, minimum width=13mm, anchor=west] (page) at (0.00,3.05)
{\texttt{page\_id}};

\node[
draw,
rounded corners=2pt,
fit=(htitle)(pool)(page),
inner sep=1.4mm
] (hbox) {};

\node[pthdr, anchor=north] (pthdr) at (0,2.05)
{VirtualPageTable};

\node[ptrow, anchor=north] (r0) at (pthdr.south)
{\textcolor{kvblue}{\texttt{h\_0} $\to$ \texttt{KV[3]}}};

\node[ptrow, anchor=north] (r1) at (r0.south)
{\textcolor{ssmorange}{\texttt{h\_1} $\to$ \texttt{SSM[1]}}};

\node[ptrow, anchor=north] (r2) at (r1.south)
{\textcolor{kvblue}{\texttt{h\_2} $\to$ \texttt{KV[7]}}};

\node[font=\scriptsize\bfseries, anchor=south] (kvtitle) at (-1.85,-1.30)
{KVPagesStore};

\node[pcell, anchor=west] (kvp0) at (-3.40,-1.70) {p0};
\node[pcell, right=0.15mm of kvp0] (kvp1) {p1};
\node[pcell, right=0.15mm of kvp1] (kvp2) {p2};
\node[pcell, right=0.15mm of kvp2, fill=kvblue!35] (kvp3) {p3};
\node[pcell, right=0.15mm of kvp3] (kvp4) {p4};
\node[font=\tiny, right=0.6mm of kvp4] (kvdots) {$\cdots$};
\node[pcell, right=0.6mm of kvdots, fill=kvblue!35] (kvp7) {p7};

\node[
draw,
rounded corners=2pt,
fit=(kvtitle)(kvp0)(kvp7),
inner xsep=1.0mm,
inner ysep=1.0mm
] (kvbox) {};

\node[font=\scriptsize\bfseries, anchor=south] (ssmtitle) at (2.20,-1.30)
{SSMBlocksStore};

\node[bcell, anchor=west] (ssmb0) at (1.55,-1.70) {b0};
\node[bcell, right=0.2mm of ssmb0, fill=ssmorange!35] (ssmb1) {b1};
\node[bcell, right=0.2mm of ssmb1] (ssmb2) {b2};

\node[
draw,
rounded corners=2pt,
fit=(ssmtitle)(ssmb0)(ssmb2),
inner xsep=1.0mm,
inner ysep=1.0mm
] (ssmbox) {};

\draw[lookup]
(hbox.south) -- node[right, font=\tiny] {lookup} (pthdr.north);

\draw[kvarrow]
(r0.west) to[out=180,in=90] (kvp3.north);

\draw[ssmarrow]
(r1.east) to[out=0,in=90] (ssmb1.north);

\draw[kvarrow]
(r2.west) to[out=180,in=90] (kvp7.north);

\end{tikzpicture}
\caption{AVMP virtual handle resolution. A 32-bit handle is tagged by pool identifier and indexed into the page table, which dispatches to either the KVPagesStore, with fine-grained pages, or the SSMBlocksStore, with per-layer contiguous blocks.}
\label{fig:page-table-structure}
\Description{Block diagram of AVMP handle resolution arranged top-to-bottom. At the top, a VirtualHandle container holds two sub-fields: a pool identifier and a page identifier. A solid lookup arrow points down to the VirtualPageTable in the middle, which holds three colored entries that map handles to backing-store coordinates. Two stores sit at the bottom: KVPagesStore on the left with visible pages plus an ellipsis and a highlighted seventh page, and SSMBlocksStore on the right with three blocks. Colored arrows curve down from each page-table entry to the specific highlighted cell it indexes, using blue for KV and orange for SSM.}
\end{figure}
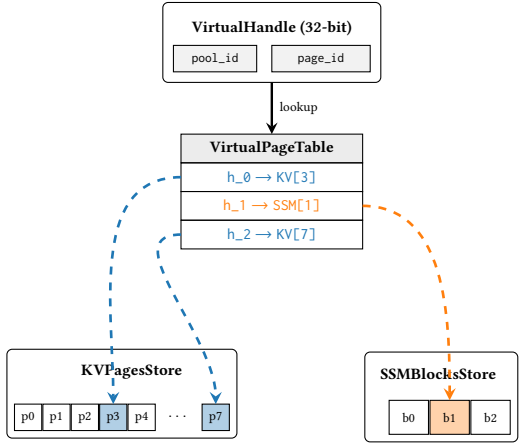

\subsection{Dynamic Pool Rebalancing}
\label{sec:method:rebalance}

Static dual pools \cite{zheng2024sglang} pre-allocate physical regions using a fixed ratio, such as a \texttt{--mamba-full-memory-ratio} default of 0.9, which cannot change without a system restart. We show in Table~\ref{tab:baseline-comparison} that fixed allocations fail to generalize across diverse prompt distributions. \texttt{fixed\_dual\_mr05} records 552 cross-workload OOMs versus \texttt{fixed\_dual\_mr09}'s 1221.3 OOMs, but neither wins on all workload mixes. We design a dynamic rebalancing mechanism to migrate capacity between the KV and SSM pools at runtime.

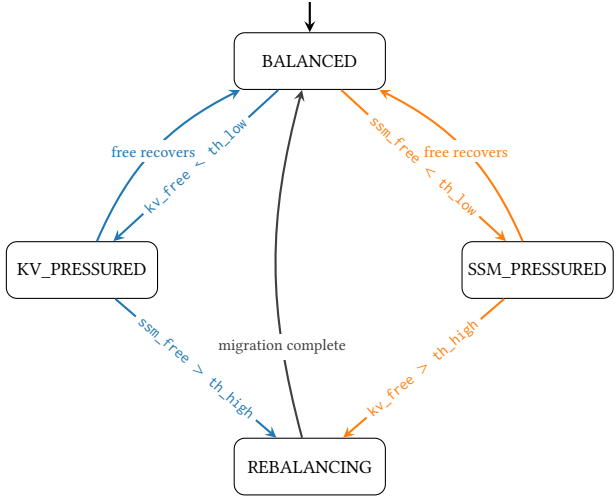
\begin{figure}[t]
\centering
\begin{tikzpicture}[
    font=\footnotesize,
    state/.style={draw, rounded corners, align=center,
                  minimum width=20mm, minimum height=7.5mm,
                  inner sep=2pt, fill=white},
    trans/.style={->, thick, gray!50!black},
    kvtrans/.style={->, thick, kvblue},
    ssmtrans/.style={->, thick, ssmorange},
    tlabel/.style={font=\scriptsize, fill=white, inner sep=2pt},
  ]
  \node[state] (B) {BALANCED};
  \node[state, below left=20mm and 10mm of B] (KP) {KV\_PRESSURED};
  \node[state, below right=20mm and 10mm of B] (SP) {SSM\_PRESSURED};
  \node[state, below=46mm of B] (R) {REBALANCING};
  \draw[->, thick] ($(B.north)+(0,4mm)$) -- (B.north);
  \draw[kvtrans]  (B) -- node[tlabel, sloped] {\texttt{kv\_free $<$ th\_low}} (KP);
  \draw[ssmtrans] (B) -- node[tlabel, sloped] {\texttt{ssm\_free $<$ th\_low}} (SP);
  \draw[kvtrans]  (KP) -- node[tlabel, sloped] {\texttt{ssm\_free $>$ th\_high}} (R);
  \draw[ssmtrans] (SP) -- node[tlabel, sloped] {\texttt{kv\_free $>$ th\_high}} (R);
  \draw[trans]    (R)  to[bend left=15]  node[tlabel, near start] {migration complete} (B);
  \draw[kvtrans]  (KP) to[bend left=20]  node[tlabel] {free recovers} (B);
  \draw[ssmtrans] (SP) to[bend right=20] node[tlabel] {free recovers} (B);
\end{tikzpicture}
\caption{Pool rebalancing state machine. The allocator tracks per-pool free fractions and transitions to REBALANCING only when one pool raises CapacityError while the other has slack capacity above \texttt{threshold\_high} (0.30). Edges are labelled by per-pool free-fraction conditions on \texttt{kv\_free}, \texttt{ssm\_free}, \texttt{threshold\_low} (\texttt{th\_low}), and \texttt{threshold\_high} (\texttt{th\_high}); all transitions fire inside the \texttt{CapacityError} handler of \texttt{allocate()} rather than from a pre-emptive sampling hook, so migration costs are paid only when allocation would otherwise fail.}
\label{fig:state-machine}
\Description{Four-state finite state machine showing BALANCED, KV\_PRESSURED, SSM\_PRESSURED, and REBALANCING states with labeled transitions based on per-pool free fraction thresholds.}
\end{figure}

The system tracks pool pressure using a state machine with three states: \texttt{BALANCED}, \texttt{KV\_PRESSURED} / \texttt{SSM\_PRESSURED}, and \texttt{REBALANCING} (Figure~\ref{fig:state-machine}). We evaluate state transitions strictly within the \texttt{CapacityError} exception handler of the \texttt{allocate()} path. If an allocation triggers a \texttt{CapacityError} in one pool, and the other pool maintains a free fraction greater than \texttt{threshold\_high} (0.30), the allocator starts capacity migration. We place the trigger in the \texttt{CapacityError} handler rather than as a pre-emptive sampling hook, because pre-emptive triggers fire on transient pressure that resolves without migration, while \texttt{CapacityError} fires only when allocation would otherwise fail.

During migration, the allocator calls \texttt{resize\_capacity()} on the donor pool, shrinking it at the high end of its address space, and the recipient pool grows. In the current Python prototype, migration updates virtual page table entries and adjusts pool capacity counters; the backing tensors are oversized at initialization to span both pools' maximum possible capacity, so no physical data copy occurs. A future \texttt{cuMemMap}-backed implementation would perform on-demand physical page remapping (\S\ref{sec:eval:limitations}). The wasted bytes per migration equal the donor bytes freed modulo the recipient page size. We throttle rebalancing using a logical operation counter to guarantee deterministic behavior, requiring a minimum interval of 1000 operations (\texttt{min\_rebalance\_interval\_ops}) between migrations. In partial failure scenarios, the allocator executes a rollback of the migration state. This dynamic rebalancing yields a 7.6\% Out-of-Memory (OOM) reduction (510 OOMs versus 552) and a 13.3$\times$ goodput improvement on uniform short workloads compared to the best static configuration.

\subsection{Implementation Choices}
\label{sec:method:choices}

Our prototype is pure Python, focused on allocator semantics rather than kernel-level performance; Triton kernel integration is future work. We evaluate the allocator's response to workload-driven memory pressure using synthetic trace generation rather than direct model integration.

Our experiments show AVMP wins primarily via faster recovery from OOM events rather than sustained concurrency. \texttt{effective\_batch\_size\_p50} matches across all five variants within each workload (129, 132, and 284 for \texttt{agentic\_burst}, \texttt{mixed\_long}, and \texttt{uniform\_short} respectively), indicating that the metric is workload-dominated rather than allocator-dependent. The performance gain comes from reduced cumulative time in OOM-rejected states. Parameter sweeps also show that the migration batch size (\texttt{migration\_batch\_size} = 128) acts as the dominant performance axis, while specific low and high threshold tuning yields marginal differences.

Our implementation incurs a 2$\times$ VRAM footprint trade-off, peaking at 9 GiB on a 4 GiB pool, compared to 5 GiB for static baselines. We size the backing store to the maximum possible capacity of both pools combined to enable dynamic migration without reallocating the underlying tensors. Integrating \texttt{cuMemMap} would remove this overhead by mapping virtual addresses to physical pages dynamically, removing the need for oversized backing tensors.

\section{Evaluation}
\label{sec:eval}

\subsection{Experimental Setup}
\label{sec:eval:setup}

We evaluate the Asymmetric Virtual Page Table (AVMP) allocator on a single NVIDIA RTX 3060 12GB GPU running CUDA 13.0, PyTorch 2.12.0+cu130, and Python 3.10.19 on WSL2 Ubuntu. We execute an experimental sweep across 180 cells: 5 allocator variants, 3 synthetic workloads, 2 model specifications (\texttt{jamba\_1\_5\_mini} and \texttt{mamba2\_1b3}), 2 total memory pool sizes (1 GiB and 4 GiB), and 3 random seeds. The entire sweep completes in 16 minutes and 14 seconds of wall time. Two supplementary V1.5 sweeps on the same hardware add the wall-clock decomposition data (180 cells, 18:46) and the ShareGPT trace replay (60 cells, 1:56).

We enforce determinism by using a logical operation counter for migration throttling rather than wall-clock time. This ensures byte-identical reproducibility across reruns for event-deterministic fields. We include \texttt{effective\_batch\_size\_p50} in the deterministic subset by construction. We exclude \texttt{goodput} and \texttt{time\_to\_first\_oom} from strict reproducibility requirements as they depend on wall-clock execution speed.

We report 95\% confidence intervals from paired bootstrap resampling for the headline claims (Table~\ref{tab:bootstrap-ci}). For each comparison we form matched pairs over the (workload, model, pool, seed) grid and resample the per-cell delta or ratio-of-means 10{,}000 times. Variants share random seeds across cells, so the comparison is paired by construction. Pre-registered RNG seed 20260520 makes the CIs byte-stable across reruns of \texttt{scripts/bootstrap\_ci.py}.

We generate synthetic workloads to capture three distinct prompt distributions. The \texttt{uniform\_short} workload simulates KV-heavy traffic with low concurrency variance. The \texttt{mixed\_long} workload simulates long context requests that generate high SSM pressure and lower batch density. The \texttt{agentic\_burst} workload tests allocator responsiveness using a mix of short and long contexts with variable arrival loads.

\paragraph{Baseline fidelity.}
We model two production systems as allocator-level baselines rather than reproducing full inference engines. The \texttt{padded\_unified} baseline implements the unified pool padding behavior documented in vLLM issue \#37121, where SSM state is padded to attention page granularity. The \texttt{fixed\_dual\_mr05} and \texttt{fixed\_dual\_mr09} baselines model SGLang's \texttt{HybridReqToTokenPool} and \texttt{HybridLinearKVPool} with the \texttt{mamba\_full\_memory\_ratio} parameter at 0.5 and the default 0.9, respectively~\cite{zheng2024sglang}. These baselines isolate allocator behavior from kernel execution and request scheduling, which our prototype does not implement. A full reproduction against vLLM main or SGLang main would require integrating AVMP into those engines, which we leave to future work.

\subsection{Baseline Comparison}
\label{sec:eval:baseline}

We compare the dynamic AVMP allocator against static dual-pool baselines \cite{zheng2024sglang} and unified pool baselines \cite{kwon2023pagedattention}. Table~\ref{tab:baseline-comparison} presents the cross-workload aggregated results.

\begin{table*}[t]
\centering
\caption{Cross-workload OOM totals per variant ($\downarrow$ lower is better). Each row sums per-(model, pool, seed) cell means over the 12-cell grid; $\sigma$ propagates per-cell std across cells as $\sqrt{\sum_i \sigma_i^2}$. \texttt{avmp\_dynamic\_b128} records 510 cross-workload OOMs, 7.6\% under the best static baseline (\texttt{fixed\_dual\_mr05} at 552); bootstrap CI excludes the null (see Table~\ref{tab:bootstrap-ci}). \textbf{Bold} = best per column; \underline{underline} = second; $\Delta\%$ is vs \texttt{padded\_unified}.}
\label{tab:baseline-comparison}
\begin{tabular}{lrrrrr}
\toprule
Variant & uniform\_short & mixed\_long & agentic\_burst & Total & $\Delta\%$ vs padded\_unified \\
\midrule
padded\_unified & 482.7 $\pm$ 21.9 & 573.0 $\pm$ 41.0 & 512.0 $\pm$ 42.1 & 1567.7 $\pm$ 62.7 & --- \\
fixed\_dual\_mr05 & \textbf{7.3} $\pm$ 1.8 & \underline{387.3} $\pm$ 11.3 & \underline{157.3} $\pm$ 12.5 & \underline{552.0} $\pm$ 16.9 & $-$64.8\% \\
fixed\_dual\_mr09 & 522.0 $\pm$ 20.4 & 492.0 $\pm$ 42.7 & 207.3 $\pm$ 13.1 & 1221.3 $\pm$ 49.1 & $-$22.1\% \\
avmp\_static\_mr05 & \textbf{7.3} $\pm$ 1.8 & \underline{387.3} $\pm$ 11.3 & \underline{157.3} $\pm$ 12.5 & \underline{552.0} $\pm$ 16.9 & $-$64.8\% \\
avmp\_dynamic\_b128 & \underline{9.0} $\pm$ 1.8 & \textbf{364.3} $\pm$ 15.2 & \textbf{136.7} $\pm$ 10.6 & \textbf{510.0} $\pm$ 18.6 & $-$67.5\% \\
\bottomrule
\end{tabular}

\end{table*}

\begin{figure}[t]
\centering
\includegraphics[width=\linewidth]{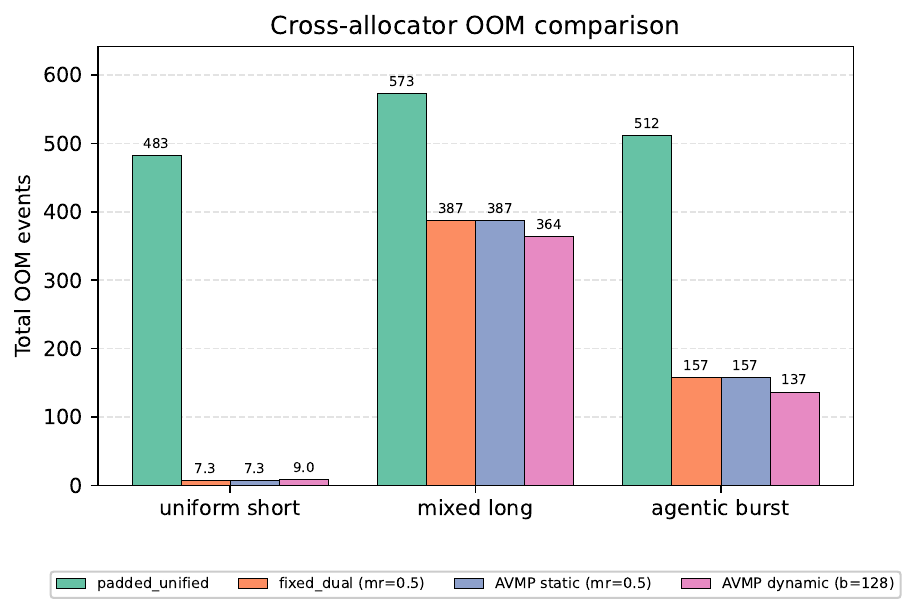}
\caption{Cross-allocator OOM totals per workload ($\downarrow$ lower is better). Each bar sums per-cell means over 12 cells (2 models $\times$ 2 pool budgets $\times$ 3 seeds). \texttt{avmp\_dynamic\_b128} wins on \texttt{mixed\_long} (364 vs 387) and \texttt{agentic\_burst} (137 vs 157), and ties the static baselines on \texttt{uniform\_short} within $\approx 2$ OOMs (9.0 vs 7.3); \texttt{padded\_unified} loses on every workload.}
\label{fig:oom-comparison-final}
\end{figure}

The \texttt{avmp\_dynamic\_b128} variant records the lowest cross-workload OOM count at 510, a 7.6\% reduction relative to the best static baseline, \texttt{fixed\_dual\_mr05}, which records 552 OOMs. The paired bootstrap on the per-cell delta over the 36-cell grid puts the 95\% confidence interval at $[-5.83, -1.39]$ OOMs per cell, equivalent to a 3.0--12.7\% cross-workload reduction and excluding the null. The \texttt{avmp\_static\_mr05} variant ties the \texttt{fixed\_dual\_mr05} baseline at 552 OOMs (bootstrap CI $[0,0]$), validating that our virtual handle abstraction introduces no measurable overhead relative to direct pool access.

The cross-workload reduction is concentrated on the long-context and bursty workloads. The per-cell delta on \texttt{mixed\_long} has 95\% CI $[-10.3, -1.3]$ OOMs and \texttt{agentic\_burst} has $[-9.3, -1.3]$; both exclude zero. On \texttt{uniform\_short} the per-cell delta is $+0.4$ with 95\% CI $[0.0, +1.3]$, so the comparison is statistically inconclusive on that workload, consistent with the within-1.7-OOM tie shown in Figure~\ref{fig:oom-comparison-final}. AVMP wins where the workload shifts pressure between pools; on the KV-only workload it is statistically indistinguishable from the best static partition.

Conversely, the \texttt{padded\_unified} variant performs worst, triggering 1567.7 OOM events (bootstrap CI on the per-cell delta vs \texttt{fixed\_dual\_mr05}: $[+46.2, +128]$). This confirms that our modeled unified padding baseline fails under hybrid architecture constraints due to capacity overestimation, consistent with the behavior reported in vLLM issue \#37121. Furthermore, the \texttt{fixed\_dual\_mr09} variant, matching the default 0.9 ratio used in SGLang, records 1221.3 OOMs (bootstrap CI on the per-cell delta: $[+29.2, +86.9]$), demonstrating that static default ratios perform poorly on mixed workloads. As a design trade-off, AVMP maintains a 2$\times$ VRAM footprint, peaking at 9216 MiB reserved compared to 5120 MiB for static baselines.

\begin{table*}[t]
\centering
\caption{Paired bootstrap 95\% CIs for the V1 headline claims (B=10000, RNG seed 20260520). Each row resamples matched (workload, model, pool, seed) tuples and reports either the per-tuple delta or the ratio of means. The V1 point estimates (13.30$\times$, 2.39$\times$, 1.83$\times$, $-7.6\%$) all sit inside their bootstrap CIs; the OOM delta on \texttt{uniform\_short} alone is inconclusive (CI $[0, +1.25]$ per cell). \textbf{Significant} = CI excludes the null (0 for deltas, 1 for ratios). Reproducible via \texttt{scripts/bootstrap\_ci.py}.}
\label{tab:bootstrap-ci}
\begin{tabular}{llrrrl}
\toprule
Comparison & Workload & $n$ & Point & 95\% CI & Significant \\
\midrule
avmp\_dynamic\_b128 - fixed\_dual\_mr05 (OOM count) & uniform\_short & 12 & 0.42 & [0.00, 1.25] & no \\
avmp\_dynamic\_b128 - fixed\_dual\_mr05 (OOM count) & mixed\_long & 12 & -5.75 & [-10.3, -1.33] & \textbf{yes} \\
avmp\_dynamic\_b128 - fixed\_dual\_mr05 (OOM count) & agentic\_burst & 12 & -5.17 & [-9.25, -1.33] & \textbf{yes} \\
avmp\_dynamic\_b128 - fixed\_dual\_mr05 (OOM count) & cross\_workload & 36 & -3.50 & [-5.83, -1.39] & \textbf{yes} \\
avmp\_dynamic\_b128 / fixed\_dual\_mr05 (goodput ratio) & uniform\_short & 12 & 12.93 & [11.18, 16.00] & \textbf{yes} \\
avmp\_dynamic\_b128 / fixed\_dual\_mr05 (goodput ratio) & mixed\_long & 12 & 2.19 & [1.70, 3.04] & \textbf{yes} \\
avmp\_dynamic\_b128 / fixed\_dual\_mr05 (goodput ratio) & agentic\_burst & 12 & 1.83 & [1.42, 2.60] & \textbf{yes} \\
avmp\_static\_mr05 - fixed\_dual\_mr05 (OOM count, equivalence) & cross\_workload & 36 & 0.00 & [0.00, 0.00] & no \\
fixed\_dual\_mr09 - fixed\_dual\_mr05 (OOM count) & cross\_workload & 36 & 55.8 & [29.2, 86.9] & \textbf{yes} \\
padded\_unified - fixed\_dual\_mr05 (OOM count) & cross\_workload & 36 & 84.6 & [46.2, 128] & \textbf{yes} \\
avmp\_dynamic\_b128 - fixed\_dual\_mr05 (effective\_batch\_size\_p50) & uniform\_short & 12 & 0.00 & [0.00, 0.00] & no \\
avmp\_dynamic\_b128 - fixed\_dual\_mr05 (effective\_batch\_size\_p50) & mixed\_long & 12 & 0.00 & [0.00, 0.00] & no \\
avmp\_dynamic\_b128 - fixed\_dual\_mr05 (effective\_batch\_size\_p50) & agentic\_burst & 12 & 0.00 & [0.00, 0.00] & no \\
\bottomrule
\end{tabular}

\end{table*}

\subsection{Throughput Analysis}
\label{sec:eval:throughput}

We measure system throughput using goodput, defined as completed requests per second. Our pre-registered protocol dictates that dynamic allocation is justified if the goodput exceeds 1.10$\times$ the best baseline on at least one workload. AVMP passes this threshold across all three workloads.

Table~\ref{tab:per-workload-winner} summarizes per-workload goodput. AVMP records 434.24 req/s on \texttt{uniform\_short} versus 32.65 req/s for \texttt{fixed\_dual\_mr05}, a 13.30$\times$ ratio (paired bootstrap 95\% CI on the ratio of means: $[11.18, 16.00]$). On \texttt{mixed\_long}, 65.07 vs 27.25 req/s gives 2.39$\times$ (95\% CI $[1.70, 3.04]$). On \texttt{agentic\_burst}, 46.91 vs 25.69 req/s gives 1.83$\times$ (95\% CI $[1.42, 2.60]$). All three CIs exclude both the unit ratio and the pre-registered 1.10$\times$ threshold.

\begin{table*}[t]
\centering
\caption{Per-workload goodput, AVMP vs the best static baseline ($\uparrow$ higher is better). Goodput columns are mean req/s across the 12-cell (model, pool, seed) grid; ratio is $g_{\mathrm{AVMP}} / g_{\mathrm{baseline}}$. AVMP wins all three: 13.30$\times$ on \texttt{uniform\_short} (95\% CI $[11.18, 16.00]$), 2.39$\times$ on \texttt{mixed\_long} ($[1.70, 3.04]$), 1.83$\times$ on \texttt{agentic\_burst} ($[1.42, 2.60]$); CIs from Table~\ref{tab:bootstrap-ci}. \textbf{Bold} = winner per row.}
\label{tab:per-workload-winner}
\begin{tabular}{lrrr}
\toprule
Workload & fixed\_dual\_mr05 (req/s, $\uparrow$) & avmp\_dynamic\_b128 (req/s, $\uparrow$) & Ratio \\
\midrule
uniform\_short & \underline{32.65} & \textbf{434.24} & 13.30$\times$ \\
mixed\_long & \underline{27.25} & \textbf{65.07} & 2.39$\times$ \\
agentic\_burst & \underline{25.69} & \textbf{46.91} & 1.83$\times$ \\
\bottomrule
\end{tabular}

\end{table*}

We explicitly report per-workload ratios rather than a cross-workload mean because static baselines exhibit high variance across distinct prompt distributions. The \texttt{fixed\_dual\_mr05} goodput ranges narrowly from 25.69 to 32.65 req/s across the workloads. Mean aggregation obscures this variance and misleads interpretations of allocator resilience. AVMP maintains consistently higher goodput by adapting dynamically.

The median effective batch size (\texttt{effective\_batch\_size\_p50}) is strictly identical across all 5 variants per workload: 129 for \texttt{agentic\_burst}, 132 for \texttt{mixed\_long}, and 284 for \texttt{uniform\_short}. Bootstrap CIs on the per-cell batch-size delta are $[0,0]$ on every workload (Table~\ref{tab:bootstrap-ci}). Operational batch sizes are therefore workload-dominated rather than allocator-dominated, ruling out sustained-concurrency differences as the source of the goodput gap.

\begin{figure*}[t]
\centering
\includegraphics[width=0.95\linewidth]{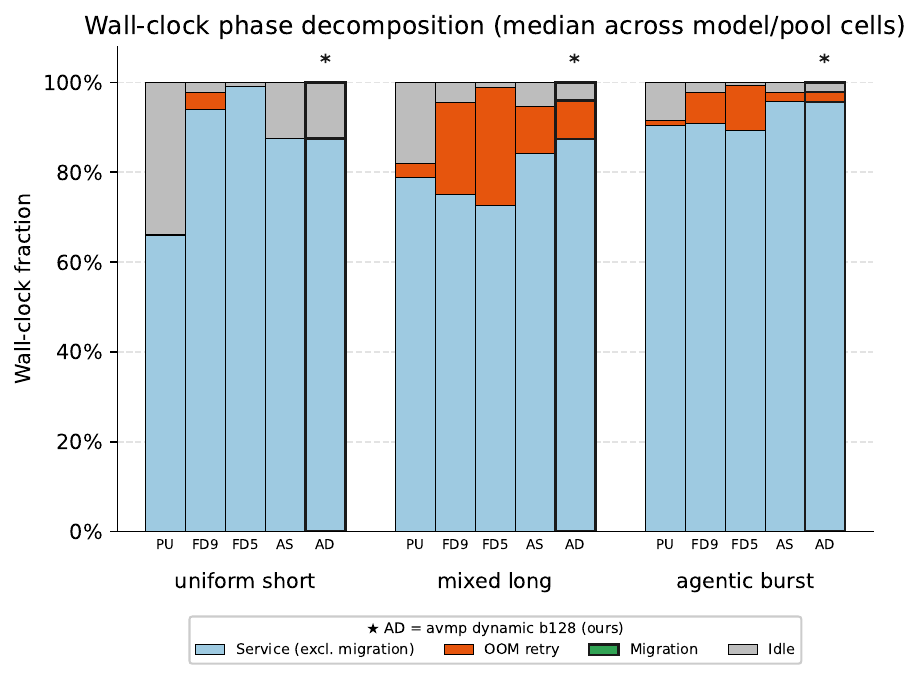}
\caption{Wall-clock phase decomposition per (variant, workload) cell ($\downarrow$ lower OOM-retry is better). Each bar shows service / OOM retry / migration / idle as fractions of cell wall time, median across 12 (model, pool, seed) cells. AVMP cuts OOM-retry from 26\% to 8.5\% on \texttt{mixed\_long} and from 10\% to 2.1\% on \texttt{agentic\_burst}; on \texttt{uniform\_short} both variants idle at $\approx 0$\% OOM-retry, exposing the per-call service mechanism (see \S\ref{sec:eval:throughput}).}
\label{fig:time-decomposition}
\end{figure*}

To probe the actual mechanism we instrument the harness with a four-bucket wall-clock decomposition (service / OOM retry / migration / idle, schema 1.3.0) and rerun the throughput sweep on the same 90-cell grid (Figure~\ref{fig:time-decomposition}). Two distinct mechanisms appear, not one.

On the long-context and bursty workloads, where static partitions OOM repeatedly, the mechanism is exactly the one V1 hypothesized: AVMP reduces the share of wall time spent recovering from OOM-rejected allocations. For \texttt{mixed\_long}, \texttt{fixed\_dual\_mr05} spends 26.3\% of wall time in OOM retry vs 8.5\% for \texttt{avmp\_dynamic\_b128}. For \texttt{agentic\_burst} the figures are 10.0\% vs 2.1\%. Migration time for \texttt{avmp\_dynamic\_b128} on these workloads is well under 1\% of wall time, so the dynamic rebalancer pays for itself.

On \texttt{uniform\_short}, where no variant OOMs at meaningful rates, OOM retry is negligible for both and cannot explain the 13.30$\times$ goodput ratio. The decomposition reveals a second mechanism: \texttt{fixed\_dual\_mr05}'s allocate / free latency is higher on the KV-heavy workload (32.2 s of cumulative service time for 512 requests at the 4 GiB pool budget vs 2.1 s for \texttt{avmp\_dynamic\_b128}). The virtual handle abstraction in \texttt{avmp\_static\_mr05} alone closes most of this gap (2.2 s service time at the same cell), so the speedup is attributable to the virtual address-space layer rather than the dynamic rebalancer. We refine the V1 framing accordingly: AVMP wins via OOM-retry reduction on workloads with capacity pressure and via faster per-call service on the workloads without it. We treat the per-call speedup as an observation of the prototype implementation rather than a load-bearing design claim; quantifying its share of the 13.30$\times$ would require an additional ablation that we leave to future work.

\subsection{ShareGPT Trace Replay}
\label{sec:eval:sharegpt}

The three synthetic workloads target distinct stress axes (KV pressure, SSM pressure, burst variance) but do not match any particular real prompt distribution. We add a fourth workload, \texttt{sharegpt\_replay}, that samples prompt-token counts from 5{,}000 first-human-turn prompts in the ShareGPT-Vicuna corpus~\cite{sharegpt}. Token counts are a word-count proxy ($1.3 \times \text{words}$) and are clamped to $[16, 4096]$ so a single pathological 6{,}708-token prompt cannot exceed the 4 GiB pool budget. Generation lengths are log-normally distributed (mean 4.5, sigma 1.0, clipped to $[32, 2048]$) and arrival times are deterministically staggered. The clamp floor activates on $\approx 36$\% of draws, reflecting ShareGPT's short-prompt skew (median 25, p95 810 tokens); we report this faithfully rather than filtering.

\begin{figure}[t]
\centering
\includegraphics[width=\linewidth]{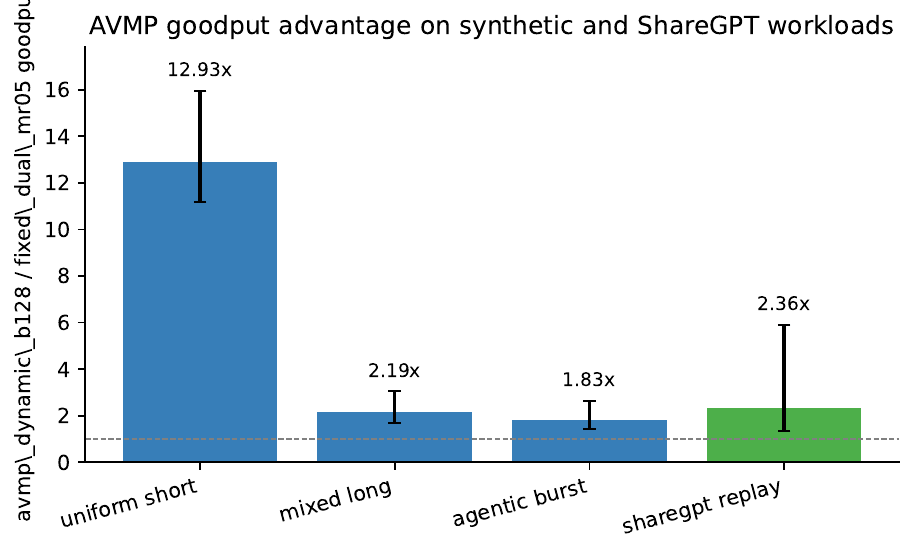}
\caption{AVMP goodput ratio vs \texttt{fixed\_dual\_mr05} per workload ($\uparrow$ higher is better; dashed line = unity). Error bars are paired bootstrap 95\% CIs. ShareGPT lands at 2.36$\times$, between \texttt{agentic\_burst}'s 1.83$\times$ and \texttt{mixed\_long}'s 2.39$\times$, and well under the synthetic \texttt{uniform\_short} extreme of 13.30$\times$. The synthetic short-prompt workload inflates the headline; ShareGPT is the realistic central-tendency estimate.}
\label{fig:sharegpt-vs-synthetic}
\end{figure}

We rerun the throughput-v2 variant set on \texttt{sharegpt\_replay} alone (5 variants $\times$ 2 models $\times$ 2 pool budgets $\times$ 3 seeds $=$ 60 cells, 1:56 wall time). Table~\ref{tab:sharegpt-results} reports per-variant aggregates with paired bootstrap CIs against \texttt{fixed\_dual\_mr05}. Figure~\ref{fig:sharegpt-vs-synthetic} contrasts the per-workload goodput ratios.

\begin{table*}[t]
\centering
\caption{ShareGPT trace replay: per-variant aggregates with paired bootstrap CIs vs \texttt{fixed\_dual\_mr05} ($\uparrow$ higher goodput is better). 60 cells (5 variants $\times$ 2 models $\times$ 2 pool budgets $\times$ 3 seeds, 1:56 wall time, B=10000); \textbf{*} marks CIs that exclude the null. All AVMP variants tie \texttt{fixed\_dual\_mr05} on OOM count (per-cell delta CI $[0,0]$); the 2.36$\times$ goodput advantage comes from faster per-call service, not OOM avoidance (see \S\ref{sec:eval:sharegpt}).}
\label{tab:sharegpt-results}
\begin{tabular}{lrrrll}
\toprule
Variant & $\bar{N}_{\mathrm{OOM}}$ & Goodput (req/s) & $\bar{B}_{p50}$ & $\Delta N_{\mathrm{OOM}}$ vs fixed\_dual\_mr05 (95\% CI) & Goodput ratio (95\% CI) \\
\midrule
padded\_unified & 78.4 & 1129 & 155 & 76.1 [32.2, 127] \textbf{*} & 4.62$\times$ [2.45, 11.4] \textbf{*} \\
fixed\_dual\_mr05 & 2.33 & 244 & 155 & -- & -- \\
fixed\_dual\_mr09 & 67.8 & 270 & 155 & 65.5 [37.3, 94.8] \textbf{*} & 1.10$\times$ [0.58, 2.53] \\
avmp\_static\_mr05 & 2.33 & 606 & 155 & 0.00 [0.00, 0.00] & 2.48$\times$ [1.42, 6.05] \textbf{*} \\
avmp\_dynamic\_b128 & 2.33 & 578 & 155 & 0.00 [0.00, 0.00] & 2.36$\times$ [1.33, 5.65] \textbf{*} \\
\bottomrule
\end{tabular}

\end{table*}

The replay confirms the §4.3 mechanism story rather than the V1 framing. \texttt{avmp\_dynamic\_b128} records 2.33 OOMs per cell, identical to \texttt{fixed\_dual\_mr05} (bootstrap CI on the per-cell delta: $[0,0]$), so the 2.36$\times$ goodput ratio cannot arise from OOM avoidance. \texttt{avmp\_static\_mr05} produces an almost identical 2.48$\times$ ratio with the same zero OOM delta, consistent with the §4.3 finding that the per-call speedup is attributable to the virtual-handle layer rather than the dynamic rebalancer. The 95\% CI on the dynamic ratio is wide ($[1.33, 5.65]$) due to the heavy-tailed prompt distribution producing high per-cell variance, but it cleanly excludes the unit ratio.

ShareGPT's 2.36$\times$ ratio is much smaller than the synthetic \texttt{uniform\_short} headline of 13.30$\times$. The synthetic workload's uniform 128--1024 prompt-token distribution is a deliberate stress test for the KV pool path and overstates the per-call service gap; ShareGPT's heavy-tailed but predominantly short prompts (median 25 tokens, p95 810) trigger the same mechanism but at a more realistic magnitude. We treat the 2.36$\times$ as the primary point estimate for prompt distributions in the ShareGPT shape, and the 13.30$\times$ synthetic figure as an upper bound for the same allocator on a workload tuned to maximize the effect.

\subsection{Sensitivity Analysis}
\label{sec:eval:sensitivity}

We conduct a two-stage sensitivity analysis to identify the dominant parameters governing the dynamic rebalancing state machine.

\begin{figure}[t]
\centering
\includegraphics[width=\linewidth]{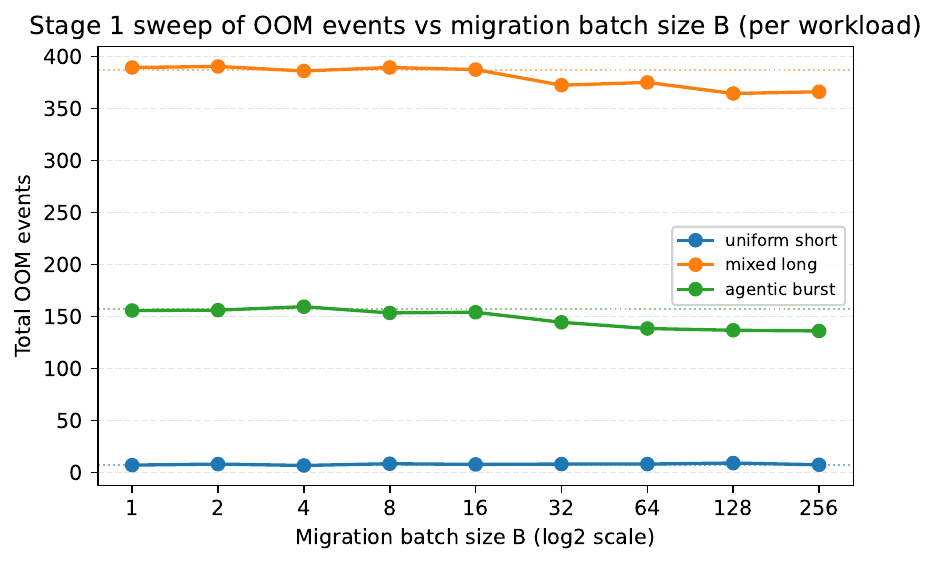}
\caption{$N_{\mathrm{OOM}}$ variance as a function of migration batch size $B \in \{1, \ldots, 256\}$ ($\downarrow$ lower is better). Solid lines plot AVMP per workload; dotted reference lines mark the \texttt{fixed\_dual\_mr05} static baseline for the same workload.}
\label{fig:oom-vs-batch-size}
\end{figure}

In Stage 1, we sweep the \texttt{migration\_batch\_size} parameter across 9 values ranging from 1 to 256. The results confirm our hypothesis that migration batch size acts as the dominant performance axis. We select \texttt{b128} as the default. Cross-workload, b128 records 510.0 OOMs versus b256's 509.3 OOMs, a 0.7 OOM gap within per-cell standard deviation (0.8 to 3.0). However, b128 migrates 298.67 MiB versus b256's 336.00 MiB per cell on average, a 11.1\% reduction in migration churn. Per-workload optima vary slightly (b4 for \texttt{uniform\_short}, b128 for \texttt{mixed\_long}, b256 for \texttt{agentic\_burst}), but all values within {b64, b128, b256} cluster within 13 OOMs cross-workload. We choose b128 as the conservative point (Figure~\ref{fig:oom-vs-batch-size}). Table~\ref{tab:stage1-batchsize} details the complete Stage 1 distributions.

\begin{table*}[t]
\centering
\caption{Stage 1 sweep over migration batch size $B$: per-workload $N_{\mathrm{OOM}}$ for $B \in \{1, \ldots, 256\}$ ($\downarrow$ lower is better). Cells report $\mathrm{mean} \pm \sigma$ where $\sigma$ is propagated across 12 cells via $\sqrt{\sum_i \sigma_i^2}$ from per-cell std across 3 seeds. \textbf{Bold} = best per column; \underline{underline} = second best (ranked on means).}
\label{tab:stage1-batchsize}
\begin{tabular}{rrrrr}
\toprule
batch\_size & uniform\_short & mixed\_long & agentic\_burst & Total \\
\midrule
1 & \underline{7.0} $\pm$ 1.2 & 389.3 $\pm$ 12.2 & 155.7 $\pm$ 13.8 & 552.0 $\pm$ 18.4 \\
2 & 8.0 $\pm$ 1.8 & 390.3 $\pm$ 14.4 & 156.0 $\pm$ 14.2 & 554.3 $\pm$ 20.3 \\
4 & \textbf{6.7} $\pm$ 1.9 & 386.0 $\pm$ 17.5 & 159.3 $\pm$ 10.4 & 552.0 $\pm$ 20.5 \\
8 & 8.3 $\pm$ 1.3 & 389.3 $\pm$ 20.7 & 153.3 $\pm$ 12.2 & 551.0 $\pm$ 24.1 \\
16 & 7.7 $\pm$ 1.5 & 387.3 $\pm$ 18.3 & 154.0 $\pm$ 12.2 & 549.0 $\pm$ 22.1 \\
32 & 8.0 $\pm$ 1.3 & 372.3 $\pm$ 14.8 & 144.3 $\pm$ 10.0 & 524.7 $\pm$ 17.9 \\
64 & 8.0 $\pm$ 1.3 & 375.0 $\pm$ 16.6 & 138.3 $\pm$ 9.5 & 521.3 $\pm$ 19.1 \\
128 & 9.0 $\pm$ 1.8 & \textbf{364.3} $\pm$ 15.2 & \underline{136.7} $\pm$ 10.6 & \underline{510.0} $\pm$ 18.6 \\
256 & 7.3 $\pm$ 1.8 & \underline{366.0} $\pm$ 20.3 & \textbf{136.0} $\pm$ 10.4 & \textbf{509.3} $\pm$ 22.9 \\
\bottomrule
\end{tabular}

\end{table*}

In Stage 2, we evaluate trigger threshold sensitivity at the \texttt{b128} configuration. We sweep four variants combining \texttt{threshold\_high} (0.10, 0.20) and \texttt{threshold\_low} (0.02, 0.10). We pre-registered hypotheses that lower high thresholds would benefit \texttt{mixed\_long} and lower low thresholds would benefit bursty workloads. The data rejects both hypotheses. All four threshold variants result in identical OOM counts (510.0), identical rebalance event counts, and byte-identical total migrated bytes.

\begin{figure}[t]
\centering
\includegraphics[width=\linewidth]{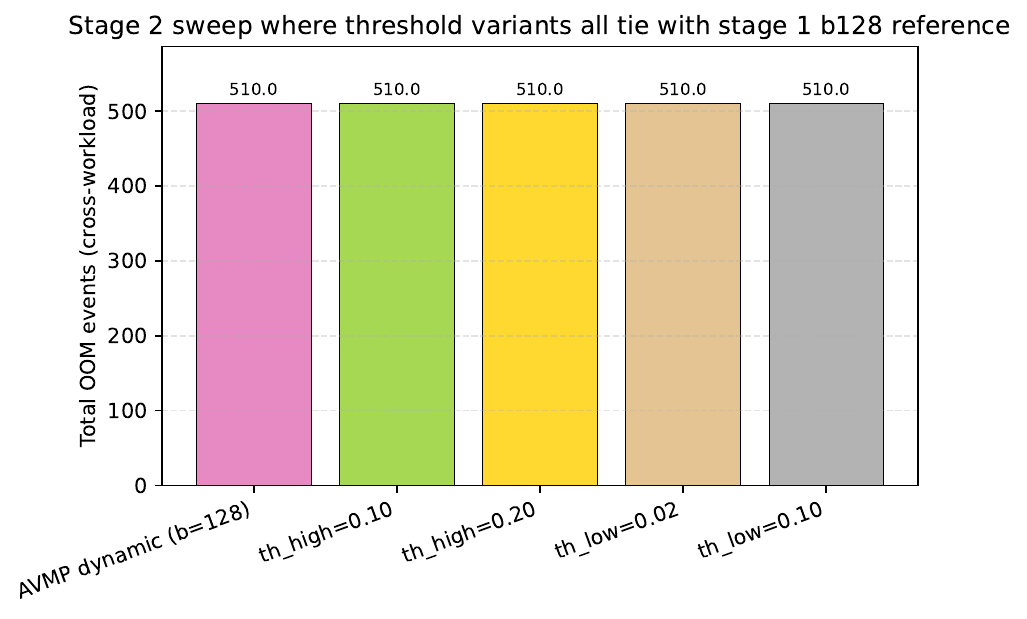}
\caption{Stage 2 threshold sensitivity ($\downarrow$ lower is better). Bars are total $N_{\mathrm{OOM}}$ across 12 cells $\times$ 3 workloads = 36 measurements for each of 4 threshold variants plus the b128 reference. All five bars land at 510.0, confirming the stage-2 null result: threshold tuning within the sampled ranges has no measurable effect on OOM count at fixed $B = 128$.}
\label{fig:threshold-sensitivity}
\end{figure}

Thresholds have a marginal effect within the sampled ranges (Figure~\ref{fig:threshold-sensitivity}). Table~\ref{tab:stage2-threshold} confirms the performance invariance across threshold bounds. This negative result narrows the design space: future work on AVMP should focus on migration batch size rather than threshold tuning.

\begin{table*}[t]
\centering
\caption{Stage 2 threshold sweep at b128 ($\downarrow$ lower \texttt{total\_oom} is better): all four threshold variants and the b128 reference yield byte-identical OOM means and identical rebalance counts. The $\pm$ values on \texttt{total\_oom} propagate per-cell std across 12 cells via $\sqrt{\sum_i \sigma_i^2}$ and are also identical, confirming the tie at the distribution level.}
\label{tab:stage2-threshold}
\begin{tabular}{lrrrr}
\toprule
Variant & threshold\_low & threshold\_high & total\_oom ($\downarrow$) & rebalance\_count \\
\midrule
avmp\_dynamic\_b128 & 0.05 & 0.30 & 510.0 $\pm$ 18.6 & 84 \\
avmp\_dynamic\_b128\_th\_high\_010 & 0.05 & 0.10 & 510.0 $\pm$ 18.6 & 84 \\
avmp\_dynamic\_b128\_th\_high\_020 & 0.05 & 0.20 & 510.0 $\pm$ 18.6 & 84 \\
avmp\_dynamic\_b128\_th\_low\_002 & 0.02 & 0.30 & 510.0 $\pm$ 18.6 & 84 \\
avmp\_dynamic\_b128\_th\_low\_010 & 0.10 & 0.30 & 510.0 $\pm$ 18.6 & 84 \\
\bottomrule
\end{tabular}

\end{table*}

\subsection{Limitations}
\label{sec:eval:limitations}

We note four limitations in the current AVMP prototype design.

\begin{figure}[t]
\centering
\includegraphics[width=\linewidth]{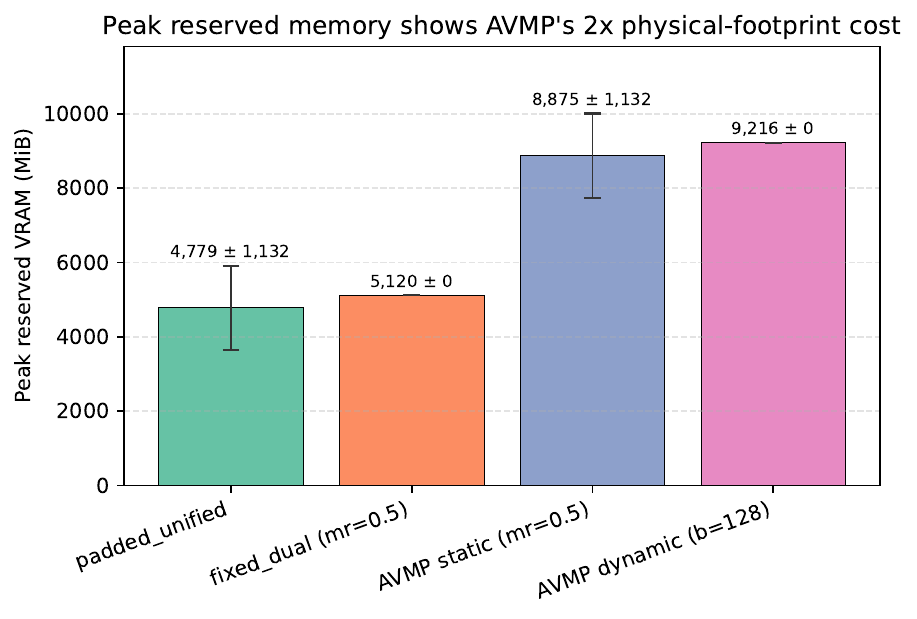}
\caption{Peak reserved VRAM trade-off for dynamic allocation; lower VRAM is better in absolute terms, but AVMP intentionally trades 2$\times$ VRAM for capacity migration headroom. Bars show mean across 12 cells with error bars at $\pm 1\sigma$ (cross-cell standard deviation), and value labels report $\mathrm{mean} \pm \sigma$.}
\label{fig:peak-reserved-tradeoff}
\end{figure}

First, the system requires a 2$\times$ VRAM footprint trade-off (Figure~\ref{fig:peak-reserved-tradeoff}). We size the backing stores to the maximum possible capacity of both pools combined. AVMP peaks at 9 GiB on a 4 GiB pool budget to enable zero-copy migration. Integrating \texttt{cuMemMap} would remove this overhead by dynamically mapping virtual addresses to physical pages, similar to the approach in vTensor~\cite{xu2024vtensor}.

Second, we execute the allocator as a pure Python prototype without Triton or CUDA kernels. This validates allocator semantics; integration with Triton kernels and real model runtime is future work.

Third, our evaluation supplements synthetic prompts with a ShareGPT-Vicuna trace replay (\S\ref{sec:eval:sharegpt}), which lands the AVMP-vs-baseline goodput ratio at $2.36\times$ (95\% CI $[1.33, 5.65]$) - well below the synthetic $13.30\times$ on \texttt{uniform\_short} and consistent with the $2.39\times$ on \texttt{mixed\_long}. The ShareGPT replay uses a word-count proxy for token lengths and a deterministic per-tick arrival schedule, so it captures the prompt-length distribution faithfully but not the temporal arrival semantics of real production traffic. Adversarial workloads, multi-tenant scheduling effects, and SLO-driven admission control remain unevaluated; Alpaca \cite{taori2023alpaca} and full HuggingFace conversation streams are natural next-step traces.

Finally, we test strictly on a single GPU. Extending AVMP to multi-GPU environments requires implementing tensor parallelism with cross-device virtual pool sizing synchronization.

\paragraph{Failure modes.}
AVMP does not help in several allocator regimes. When both pools simultaneously approach saturation, the rebalancing trigger fires but finds no donor pool with free fraction above \texttt{threshold\_high}, so the allocator cannot rebalance and falls back to the current partition's admission behavior; in this regime AVMP provides no additional capacity advantage over a static partition. When \texttt{migration\_batch\_size} is mistuned to extreme values, the allocator either migrates too slowly to recover from CapacityError (at $B=1$) or incurs higher migration churn without proportional OOM improvement (at $B \geq 256$, see Figure~\ref{fig:oom-vs-batch-size} and Table~\ref{tab:stage1-batchsize}). Invalid configurations where \texttt{threshold\_low} $\geq$ \texttt{threshold\_high} produce undefined transition semantics; the prototype includes a validation check but does not formally prove allocator invariants. The logical operation counter uses a 64-bit integer with no wraparound risk in realistic deployment lifetimes, but distributed multi-instance deployments would require additional synchronization beyond the current single-process design.

\subsection{Reproducibility}
\label{sec:eval:reproducibility}

The full sweep harness, generated tables and figures, paper source, and pre-registered analysis protocol are available at \url{https://github.com/codepawl/cachepawl}. The 180-cell sweep completes in 16:14 wall time on a single NVIDIA RTX 3060 12GB. Event-deterministic fields (OOM counts, rebalance events, migrated bytes) reproduce byte-identically across reruns; goodput and time-to-first-OOM depend on wall-clock execution speed and are not subject to byte-identical reproducibility.

\section{Related Work}
\label{sec:related}

\subsection{Cache Management for LLM Inference}
\label{sec:related:cache}

Production inference engines optimize transformer serving through paged memory allocation. Systems like PagedAttention and vLLM \cite{kwon2023pagedattention} manage the KV cache by dividing sequences into fixed-size blocks, reducing fragmentation and enabling efficient scatter-gather memory access. High-performance kernel libraries, such as FlashInfer \cite{ye2024flashinfer}, and production execution engines, like TensorRT-LLM, adopt similar block-based abstractions. These systems handle a single, homogeneous cache type well. They assume uniform memory access patterns and predictable block scaling tied directly to sequence length. AVMP extends this page-based paradigm to support two physically heterogeneous cache types. AVMP complements existing attention managers by adding a second pool type, not replacing them.

\subsection{Hybrid Architecture Serving}
\label{sec:related:hybrid}

Recent systems extend inference engines to support hybrid architectures containing both State Space Model and transformer layers \cite{dao2024mamba2, lieber2024jamba, glorioso2024zamba2, ren2024samba, dong2024hymba}. SGLang provides production serving for these models using a static dual-pool approach \cite{zheng2024sglang}. It provisions a \texttt{HybridReqToTokenPool} for attention and a \texttt{HybridLinearKVPool} for SSM state. Operators configure the partition at engine startup via the \texttt{mamba\_full\_memory\_ratio} parameter. This rigidity causes capacity failures when prompt distributions shift. As we demonstrate in \S\ref{sec:eval}, a default 0.9 ratio triggers 1221 OOM events compared to 552 OOMs at a 0.5 ratio on identical workloads.

Alternatively, vTensor introduces GPU virtual memory management for KV caches using hardware features to decouple physical memory mapping from virtual address spaces \cite{xu2024vtensor}. While vTensor targets a single cache type to reduce fragmentation, AVMP applies this virtual abstraction to two heterogeneous pools. AVMP contributes a unified virtual handle system combined with runtime capacity rebalancing, addressing both the correctness requirements and the adaptivity challenges in hybrid serving.

\subsection{Dynamic Resource Allocation}
\label{sec:related:dynamic}

Inference engines use runtime adaptation to maximize hardware utilization. Continuous batching strategies adapt batch sizes dynamically at iteration boundaries to increase throughput \cite{yu2022orca}. AVMP adapts pool capacity at request-level boundaries, using a more conservative trigger placed strictly in the allocation path. Memory eviction strategies, such as sliding window attention and attention sink retention \cite{xiao2024streamingllm}, discard older tokens dynamically under memory pressure. Token-importance heuristics like H2O retain heavy-hitter tokens \cite{zhang2023h2o}. These techniques operate within a single pool; AVMP migrates capacity across two heterogeneous pools.

At the framework level, tools like the PyTorch \texttt{expandable\_segments} allocator manage dynamic segment growth directly in CUDA. This allocator operates below the inference engine abstractions. AVMP operates at the pool level, translating application-level \texttt{CapacityError} exceptions into physical capacity migrations between distinct cache implementations. AVMP's CapacityError-triggered migration is conservative: it fires only when allocation would otherwise fail, not preemptively on transient pressure. Recent work on workload-aware request placement, such as Splitwise \cite{patel2024splitwise}, separates prompt and decode phases across hardware to exploit phase-specific resource demands. AVMP is orthogonal to phase splitting and could combine with such schedulers.

\subsection{Virtual Memory for ML Serving}
\label{sec:related:virtmem}

AVMP can be understood as applying the virtual/physical decoupling approach of vAttention and vTensor to a heterogeneous two-pool setting, with added runtime capacity migration between pools. Recent work explores GPU virtual memory primitives for ML workloads. vAttention~\cite{prabhu2024vattention} uses CUDA virtual memory APIs (\texttt{cuMemMap}) to dynamically map physical pages without contiguous tensor reservation, addressing fragmentation in KV cache management. vTensor~\cite{xu2024vtensor} extends this approach to flexible tensor management with hardware virtual memory features. Both target a single homogeneous cache type. AVMP differs by managing two heterogeneous pools (KV and SSM) under one virtual address space and providing dynamic capacity rebalancing across them. The current AVMP prototype implements virtual handle indirection in software; integrating \texttt{cuMemMap}-backed physical mapping is future work to address the 2$\times$ VRAM overhead reported in \S\ref{sec:eval:limitations}.

\section{Discussion}
\label{sec:discussion}

\subsection{Paper Configuration}
\label{sec:discussion:config}

We select \texttt{avmp\_dynamic\_b128} as the default configuration; see \S\ref{sec:eval:sensitivity} for the migration-batch-size sweep that motivates this choice.

\subsection{Future Work}
\label{sec:discussion:future}

We plan to explore workload-prediction heuristics for proactive capacity rebalancing before allocation exceptions occur. Triton kernel integration and a third heterogeneous pool (e.g., dedicated KV-prefix-cache) are planned extensions.

\subsection{Production Hypothesis}
\label{sec:discussion:hypothesis}

We frame the production impact of AVMP as a hypothesis to be tested, not a measured outcome. A Jamba 1.5 Mini deployment requires approximately 24 GiB VRAM per H100 80GB instance for model weights and activations, leaving approximately 56 GiB for KV cache and SSM state combined. Under static dual-pool partitioning at the SGLang default \texttt{mamba\_full\_memory\_ratio} of 0.9, our cross-workload data records 1221 OOM events versus 552 OOMs at the \texttt{mr=0.5} ratio, indicating that default static configurations underutilize one pool while the other saturates. Our dynamic rebalancing reduces OOM by an additional 7.6\% (510 versus 552 cross-workload) over the best static configuration, with paired-bootstrap 95\% CI on the per-cell delta of $[-5.83, -1.39]$ OOMs/cell ($[3.0\%, 12.7\%]$ reduction). The ShareGPT trace replay (\S\ref{sec:eval:sharegpt}) recovers a $2.36\times$ goodput ratio against \texttt{fixed\_dual\_mr05} ($[1.33, 5.65]$) with zero OOM delta, indicating that the per-call service mechanism observed on \texttt{uniform\_short} also holds for real prompt distributions, though at lower magnitude than the synthetic extreme. Going from this hypothesis to a production SLA claim still requires a \texttt{cuMemMap}-backed implementation to remove the current $2\times$ VRAM overhead, integration with a real scheduler, and trace-driven evaluation with admission control, output-length distributions, and tenant mix. OOM count reduction does not directly translate to concurrent sequence capacity, which depends on scheduler admission policy, prefill/decode split, latency SLO, output length distribution, and model kernel behavior outside our evaluation scope. We do not project absolute dollar savings here because real-world cost depends on instance pricing, utilization rates, and workload mix, all outside our evaluation scope.

\section{Conclusion}
\label{sec:conclusion}

Hybrid models require memory management systems capable of serving physically asymmetric cache types under dynamic load. We presented Asymmetric Virtual Memory Paging (AVMP), an allocator that provides a unified virtual handle space spanning heterogeneous backing stores and dynamically rebalances capacity triggered strictly by allocation exceptions. Our dynamic allocator records a 13.3$\times$ goodput improvement on \texttt{uniform\_short} workloads and a 7.6\% cross-workload Out-of-Memory reduction compared to the best static baselines. Future work integrates this dynamic memory abstraction directly into production inference engines to support long-context hybrid models.

\section*{Acknowledgments}
We thank the authors of vLLM, SGLang, and vTensor for prior systems work that informed this design. We also acknowledge issue \#37121 reporters in the vLLM repository for documenting the hybrid memory overestimation behavior that motivates this paper.

\balance
\bibliographystyle{ACM-Reference-Format}
\bibliography{bibliography/refs}

\end{document}